\documentclass[11pt,a4paper]{article}
\usepackage[hyperref]{naaclhlt2018}
\usepackage{times}
\usepackage{latexsym}
\usepackage{amsmath}
\usepackage{enumitem}
\usepackage{multirow}
\usepackage{graphicx}
\usepackage{textcomp}
\usepackage{url}
\usepackage{booktabs}

\usepackage{url}
\usepackage{xcolor}

\newcommand{\mycomment}[3]{}

\newcommand{\ignore}[1]{}

\aclfinalcopy 
\def\aclpaperid{608} 

\title{Identifying Semantic Divergences in Parallel Text without Annotations}

\author{
  Yogarshi Vyas \and Xing Niu \and  Marine Carpuat \\
 Department of Computer Science  \\
University of Maryland \\
College Park, MD 20742, USA\\
  {\tt yogarshi@cs.umd.edu, xingniu@cs.umd.edu, marine@cs.umd.edu} \\}

\date{}

\begin{document}
\maketitle
\begin{abstract}
	Recognizing that even correct translations are not always semantically equivalent, we automatically detect meaning divergences in parallel sentence pairs with a deep neural model of bilingual semantic similarity which can be trained for any parallel corpus without any  manual annotation. We 
	show that our semantic model detects divergences more accurately than models based on surface features derived from word alignments, and that these divergences matter for 
	neural machine translation.
\end{abstract}

\section{Introduction}

Parallel sentence pairs are sentences that are translations of each other, and are therefore often assumed to convey the same meaning in the source and target language. 
Occasional mismatches between source and target have been primarily viewed as alignment noise \citep{GoutteCarpuatFoster2012} due to imperfect sentence alignment tools in parallel corpora drawn from translated texts \citep{Tiedemann2011,XuYvon2016}, or the noisy process of extracting parallel segments from non-parallel corpora \cite{FungCheung2004,MunteanuMarcu2005}.

In contrast, we view translation as a process that inherently introduces  meaning mismatches, so that even correctly aligned sentence pairs are not necessarily semantically equivalent. This can happen for many reasons: translation lexical choice often involves selecting between near synonyms that introduce language-specific nuances \citep{Hirst1995}, typological divergences lead to structural mismatches \citep{Dorr1994}, differences in discourse organization can make it impossible to find one-to-one sentence alignments  \citep{LiCarpuatNenkova2014}. Cross-linguistic variations in other discourse phenomena such as coreference, discourse relation and modality \cite{LapishnovaKoltunski2015} compounded with translation effects that distinguish ``translationese'' from original text \citep{KoppelOrdan2011} might also lead to meaning mismatches between source and target.

In this paper, we aim to provide empirical evidence that semantic divergences exist in parallel corpora and matter for downstream applications. This requires an automatic method to distinguish semantically equivalent sentence pairs from semantically divergent pairs, so that parallel corpora can be used more judiciously in downstream cross-lingual NLP applications. We propose a semantic model to automatically detect whether a sentence pair is  semantically divergent (Section \ref{sec:model}). While prior work relied on surface cues to detect mis-aligments, our approach focuses on comparing the meaning of words and overlapping text spans using bilingual word embeddings \citep{LuongPhamManning2015} and a deep convolutional neural network \citep{HeLin2016}. Crucially, training this model requires no manual annotation. Noisy supervision is obtained automatically borrowing techniques developed for parallel sentence extraction \cite{MunteanuMarcu2005}. Our model can thus easily be trained to detect semantic divergences in any parallel corpus.

We extensively evaluate our semantically-motivated models on intrinsic and extrinsic tasks: detection of divergent examples in two parallel English-French data sets (Section \ref{sec:intrinsic}), and data selection for English-French and Vietnamese-English machine translation (MT) (Section \ref{sec:extrinsic}).
The semantic models significantly outperform other methods on the intrinsic  task, and help select data to train neural machine translation faster with no loss in translation quality. 
Taken together, these results provide empirical evidence that sentence-alignment does not necessarily imply semantic equivalence, and that this distinction matters in practice for a downstream NLP application.

\section{Background}

\paragraph{Translation Divergences} We use the term semantic divergences to refer to bilingual sentence pairs, including translations, that do not have the same meaning. These differ from \textit{typological divergences}, which have been defined as structural differences between sentences that convey the same meaning \cite{Dorr1994,HabashDorr2002}, and reflect the fact that languages do not encode the same information in the same way.

\paragraph{Non-Parallel Corpora}  Mismatches in bilingual sentence pairs have previously been studied to extract parallel segments from non-parallel corpora, and  augment MT training data \cite[\em inter alia]{FungCheung2004,MunteanuMarcu2005, MunteanuMarcu2006,AbduI-RaufSchwenk2009,SmithQuirkToutanova2010,RiesaMarcu2012}.  Methods for parallel sentence extraction rely primarily on surface features based on translation lexicons and word alignment patterns \cite{MunteanuMarcu2005,MunteanuMarcu2006}. Similar features have proved to be useful for the related task of translation quality estimation \cite{SpeciaRajTurchi2010,SpeciaLogachevaScarton2016}, which aims to detect divergences introduced by MT errors, rather than human translation. Recently, sentence embeddings have also been used to detect parallelism  \cite{Espana-BonetVargaBarron-CedenovanGenabith2017,SchwenkDouze2017}. Although embeddings capture semantic generalizations, these models are trained with neural MT objectives, which do not distinguish semantically equivalent segments from divergent parallel segments.

\paragraph{Cross-Lingual Sentence Semantics} Cross-lingual semantic textual similarity \cite{AgirreBaneaCerDiabGonzalez-AgirreMihalceaRigauWiebe2016} and cross-lingual textual entailment \cite{NegriMehdad2010,NegriMarchettiMehdadBentivogliGiampiccolo2012,NegriMarchettiMehdadBentivogliGiampiccolo2013} seek to characterize semantic relations between sentences in two different languages beyond translation equivalence, and are therefore directly relevant to our goal. While the human judgments obtained for each task differ, they all take inputs of the same form (two segments in different languages) and output a prediction that can be interpreted as indicating whether they are equivalent in meaning or not. Models share core intuitions, relying either on MT to transfer the cross-lingual task into its monolingual equivalent \cite{JimenezBecerraGelbukh2012,ZhaoLanNiu2013}, or on features derived from MT components such as translation dictionaries and word alignments \cite{TurchiNegri2013,LoGoutteSimard2016}. Training requires manually annotated examples, either bilingual, or monolingual when using MT for language transfer.

\paragraph{Impact of mismatched sentence pairs on MT} Prior MT work has focused on noise in sentence alignment rather than semantic divergence. \newcite{GoutteCarpuatFoster2012} show that phrase-based systems are remarkably robust to noise in parallel segments. When introducing noise by permuting the target side of parallel pairs, as many as 30\% of training examples had to be permuted to degrade BLEU significantly. While such artificial noise does not necessarily capture naturally occurring divergences, there is evidence that data cleaning to remove real noise can benefit MT in low-resource settings \cite{MatthewsAmmarBhatiaFeelyHannemanSchlingerSwayamdiptaTsvetkovLavieDyer2014}.

Neural MT quality appears to be more sensitive to the nature of training examples than phrase-based systems. \newcite{ChenKuhnFosterCherryHuang2016} suggest that neural MT systems are sensitive to sentence pair permutations in domain adaptation settings. \newcite{BelinkovBisk2017} demonstrate the brittleness of character-level neural MT when exposed to synthetic noise (random permutations of words and characters) as well as natural human errors.  Concurrently with our work, \newcite{HassanAueChenChowdharyClarkFedermannHuangJunczys-DowmuntLewisLiLiuLiuLuoMenezesQinSeideTanTianWuWuEtAl2018} claim that even small amounts of noise can have adverse effects on neural MT models, as they tend to assign high probabilities to rare events. They filter out noise  and select relevant in-domain examples jointly, using similarities between sentence embeddings obtained from the encoder of a bidirectional neural MT system trained on clean in-domain data. In contrast, we  detect semantic divergence with dedicated models that require only 5000 parallel examples (see Section~\ref{sec:intrinsic}).

This work builds on our initial study of semantic divergences \cite{CarpuatVyasNiu2017}, where we provide a framework for evaluating the impact of meaning mismatches in parallel segments on MT via data selection: we show that filtering out the most divergent segments in a training corpus improves translation quality. However, we previously detect mismatches using a cross-lingual entailment classifier, which is based on surface features only, and requires manually annotated training examples \cite{NegriMarchettiMehdadBentivogliGiampiccolo2012,NegriMarchettiMehdadBentivogliGiampiccolo2013}. In this paper, we detect divergences using a semantically-motivated model that can be trained given any existing parallel corpus without manual intervention.

\section{Approach}
\label{sec:model}

We introduce our approach to detecting divergence in parallel sentences, with the goal of (1) detecting differences ranging from large mismatches to subtle nuances, (2) without manual annotation.

\paragraph{Cross-Lingual Semantic Similarity Model} We address the first requirement using a neural model that compares the meaning of sentences using a range of granularities. We repurpose the Very Deep Pairwise Interaction (VDPWI) model, which has been previously been used to detect semantic textual similarity (STS) between English sentence pairs \cite{HeLin2016}. It achieved competitive performance on data from the STS 2014 shared task \cite{AgirreBaneaCardieCerDiabGonzalez-AgirreGuoMihalceaRigauWiebe2014}, and outperformed previous approaches on sentence classification tasks \cite{HeGimpelLin2015,TaiSocherManning2015}, with fewer parameters, faster training, and without requiring expensive external resources such as WordNet.

The VDPWI model was designed for comparing the meaning of sentences in the same language, based not only on word-to-word similarity comparisons, but also on comparisons between overlapping spans of the two sentences, as learned by a deep convolutional neural network.  We adapt the model to our cross-lingual task by initializing it with bilingual embeddings. To the best of our knowledge, this is the first time this model has been used for cross-lingual tasks in such a way. We give a brief overview of the resulting model here and refer the reader to the original paper for details. Given sentences $e$ and $f$ , VDPWI models the semantic similarity between them using a pipeline consisting of five components:

\begin{enumerate}[noitemsep]
	\item \textbf{Bilingual word embeddings}: Each word in $e$ and $f$ is represented as a vector using pre-trained, bilingual embeddings. 
	
	\item \textbf{BiLSTM for contextualizing words}: Contextualized representations for words in $e$ and $f$ are obtained by choosing the output vectors at each time step obtained by running a bidirectional LSTM \cite{SchusterPaliwal1997} on each sentence.
	
	\item \textbf{Word similarity cube}: The contextualized representations are used to calculate various similarity scores between each word in \textit{e} with each word in \textit{f}. Each score thus forms a matrix and all such matrices are stacked to form a \textit{similarity cube} tensor. 
	
	\item \textbf{Similarity focus layer}: The similarity cube is fed to a similarity focus layer that re-weights the similarities in the cube to focus on highly similar word pairs, by decreasing the weights of pairs which are less similar. This output is called the \textit{focus cube}.
	
	\item\textbf{Deep convolutional network}: The focus cube is treated as an ``image''  and passed to a deep neural network, the likes of which have been used to detect patterns in images. The network consists of repeating convolution and pooling layers. Each repetition consists of a spatial convolutional layer, a Rectified Linear Unit \cite{NairHinton2010}, and a max pooling layer, followed by fully connected layers, and a softmax to obtain the final output.
\end{enumerate}

The entire architecture is trained end-to-end to minimize the Kullback-Leibler divergence \cite{Kullback1959} between the output similarity score and gold similarity score.

\paragraph{Noisy Synthetic Supervision} How can we obtain gold similarity scores as supervision for our task? We automatically construct examples of semantically divergent and equivalent sentences as follows. Since a large number of parallel sentence pairs are semantically equivalent, we use parallel sentences as positive examples. Synthetic negative examples are generated following the protocol introduced by \newcite{MunteanuMarcu2005} to distinguish parallel from non-parallel segments. Specifically, candidate negative examples are generated starting from the positive examples $\{(e_i,f_i) ~\forall i \}$ and taking the Cartesian product of the two sides of the positive examples$\{(e_i,f_j) \forall i,j ~\text{s.t.}~ i\neq j\}$. This candidate set is filtered to ensure that negative examples are not too easy to identify: we only retain pairs that are close to each other in length (a length ratio of at most 1:2), and have enough words (at least half) which have a translation in the other sentence according to a bilingual dictionary derived from automatic word alignments.
 
This process yields positive and negative examples that are a noisy source of supervision for our task, as some of the positive examples might not be fully equivalent in meaning. However, experiments will show that, in aggregate, they provide a useful signal for the VDPWI model to learn to detect semantic distinctions (Section~\ref{sec:intrinsic}).

\section{Crowdsourcing Divergence Judgments}
\label{sec:manual}

\begin{table*}[t]\footnotesize
	\begin{tabular}{llp{14cm}}
		\toprule
		& & \textbf{Equivalent with High Agreement ($n=5$)} \\
		\midrule
		subs &  en &  the epidemic took my wife, my stepson. \\
		& fr & l'\'epid\'emie a touch\'e ma femme, mon beau-fils. \\
		& gl & the epidemic touched my wife, my stepson. \\
		\midrule	
		& & \textbf{Equivalent with Low Agreement ($n=3$)}  \\ 
		\midrule
		cc & en &   cancellation policy: if cancelled up to 28 days before date of arrival, no fee will be charged. \\
		& fr & conditions d'annulation : en cas d'annulation jusqu'\`a 28 jours avant la date d'arriv\'ee, l'h\^otel ne pr\'el\`eve pas de frais sur la carte de
		cr\'edit fournie. \\
		& gl & cancellation conditions: in case of cancellation up to 28 days before arrival date, the hotel does not charge fees from the credit card given.\\

		\midrule
		& & \textbf{Divergent with Low Agreement ($n=3$)}\\
	\midrule
		cc & en  &    what does it mean when food is ``low in ash'' or ``low in magnesium''? \\
		& fr &     quels sont les avantages d’une nourriture ``r\'eduite en cendres'' et ``faible en magn\'esium'' ? \\
		& gl & what are the advantages of a food ``low in ash'' or ``low in magnesium''?\\
		\midrule
		& & \textbf{Divergent with High Agreement ($n=5$)}\\
		\midrule
		subs & 	en &     rabbit? if i told you it was a chicken, you wouldn't know the difference.\\
		& fr &        vous croirez manger du poulet. \\
		& gl & you think eat chicken \\ 
		\bottomrule
	\end{tabular}
	\caption{Randomly selected sentence pairs (English (en), French (fr) and gloss of French (gl)) annotated as divergent or equivalent, with high and low degrees of agreement between the 5 annotators. Examples are taken either from the OpenSubtitles (subs) or Common Crawl (cc) corpus.}
	\label{tab:examples}
	\vspace{-0.15in}
\end{table*}

We crowdsource annotations of English-French sentence pairs to construct test beds for evaluating our models, and also to assess how frequent semantic divergences are in parallel corpora.

\paragraph{Data Selection} We draw examples for annotation randomly from two English-French corpora, using a resource-rich and well-studied language pair, and for which bilingual annotators can easily be found. The \textbf{OpenSubtitles corpus} contains 33M sentence pairs based on translations of movie subtitles. The sentence pairs are expected to not be completely parallel given the many constraints imposed on translations that should fit on a screen and be synchronized with a movie \cite{Tiedemann2007,LisonTiedemann2016}, and the use of more informal registers which might require frequent non-literal translations of figurative language.  The \textbf{Common Crawl corpus} contains sentence-aligned parallel documents automatically mined from the Internet. Parallel documents are discovered using e.g., URL containing language code patterns, and sentences are automatically aligned after structural cleaning of HTML. The resulting corpus of 3M sentence pairs is noisy, yet extremely useful to improve translation quality for multiple language pairs and domains \cite{SmithSaint-AmandPlamadaKoehnCallison-BurchLopez2013}.

\paragraph{Annotation Protocol} Divergence annotations are obtained via Crowdflower.\footnote{\url{http://crowdflower.com}} Since this task requires good command of both French and English, we rely on a combination of strategies to obtain good quality annotations, including Crowdflower's internal worker proficiency ratings, geo-restriction, reference annotations by a bilingual speaker in our lab, and instructions that alternate between the two languages \cite{AgirreBaneaCerDiabGonzalez-AgirreMihalceaRigauWiebe2016}. 

Annotators are shown an English-French sentence pair, and asked whether they agree or disagree with the statement ``the French and English text convey the same information.'' We do not use the term ``divergent'', and let the annotators' intuitions about what constitutes the same take precedence. We set up two distinct annotation tasks, one for each corpus, so that workers only see examples sampled from a given corpus in a given job. Each example is shown to five distinct annotators.

\paragraph{Annotation Analysis} Forcing an assignment of divergent or equivalent labels by majority vote yields 
43.6\% divergent examples in OpenSubtitles, and 
38.4\% in Common Crawl. 
Fleiss' Kappa indicates moderate agreement between annotators (0.41 for OpenSubtitles and 0.49 for Common Crawl). This suggests that the annotation protocol can be improved, perhaps by using graded judgments as in Semantic Textual Similarity tasks \cite{AgirreBaneaCerDiabGonzalez-AgirreMihalceaRigauWiebe2016}, or for sentence alignment confidence evaluation \citep{XuYvon2016}. 
Current annotations are nevertheless useful, and different degrees of agreement reveal nuances in the nature of divergences (Table~\ref{tab:examples}). Examples labeled as divergent with high confidence (lowest block of the table) are either unrelated or one language misses significant information that is present in the other. Examples labeled divergent with lower confidence contain more subtle differences (e.g., ``what does it mean'' in English vs. ``what are the advantages'' in French). 

\section{Divergence Detection Evaluation}
\label{sec:intrinsic}
Using the two test sets obtained above, we can evaluate the accuracy of our cross-lingual semantic divergence detector, and compare it against a diverse set of baselines in controlled settings. We test our hypothesis that semantic divergences are more than alignment mismatches by comparing the semantic divergence detector with models that capture mis-alignment  (Section~\ref{subsubsec:basemm}) or translation (Section~\ref{subsubsec:basemt}). Then, we compare the deep convolutional architecture of the semantic divergence model, with a much simpler model that directly compares bilingual sentence embeddings (Section \ref{subsubsec:baseembed}). Finally, we compare our model trained on synthetic examples with a supervised classifier used in prior work to predict finer-grained textual entailment categories based on manually created training examples (Section \ref{subsubsec:baseclte}). Except for the entailment classifier which uses external resources, all models are trained on the exact same parallel corpora (OpenSubtitles or CommonCrawl for evaluating on the corresponding test bed.)

\subsection{Neural Semantic Divergence Detection}

\paragraph{Model and Training Settings}  We use the publicly available implementation of the VDPWI model.\footnote{\url{https://github.com/castorini/VDPWI-NN-Torch}} We initialize with 200 dimensional BiVec French-English word embeddings \cite{LuongPhamManning2015}, trained on the parallel corpus from which our test set is drawn. We use the default setting for all other VDPWI parameters. The model is trained for 25 epochs and the model that achieves the best Pearson correlation coefficient on the validation set is chosen. At test time, VDPWI outputs a score $\in [0, 1]$, where a higher value indicates less divergence. We tune a threshold on development data to convert the real-valued score to binary predictions. 

\paragraph{Synthetic Data Generation} The synthetic training data is constructed using a random sample of 5000 sentences from the training parallel corpus as positive examples. We generate negative examples automatically as described in Section \ref{sec:model}, and sample a subset to maintain a 1:5 ratio of positive to negative examples.\footnote{We experimented with other ratios and found that the results only slightly degraded while using a more balanced ratio (1:1, 1:2), but severely degraded with a skewed ratio (1:9).}

\subsection{Parallel vs. Non-parallel Classifier} 
\label{subsubsec:basemm}

Are divergences observed in parallel corpora more than alignment errors?  To answer this question, we reimplement the model proposed by \newcite{MunteanuMarcu2005}. It discriminates parallel pairs from non-parallel pairs in comparable corpora using a supervised linear classifier 
with the following features for each sentence pair $(e,f)$:
\begin{itemize}[noitemsep]
	\item Length features: $\lvert f \rvert$, $\lvert e \rvert$,$\frac{\lvert f \rvert}{\lvert e \rvert}$, and $\frac{\lvert e \rvert}{\lvert f \rvert}$
	\item Alignment features (for each of $e$ and $f$):\footnote{Alignments are obtained using IBM Model 2 trained in each direction, combined with \texttt{union}, \texttt{intersection}, and \texttt{grow-diag-final-and} heuristics.}
	\begin{itemize}[noitemsep]
		\item Count and ratio of unaligned words
		\item Top three largest fertilities
		\item Longest contiguous unaligned and aligned sequence lengths
	\end{itemize}
	\item Dictionary features:\footnote{The bilingual dictionary used to extract features is constructed using word alignments from a random sample of a million sentences from the training parallel corpus.}
	 fraction of words in $e$ that have a translation in $f$ and vice-versa.
\end{itemize}

Training uses the exact same synthetic examples as the semantic divergence model (Section \ref{sec:model}). 

\subsection{Neural MT}
\label{subsubsec:basemt}

If divergent examples are nothing more than bad translations, a neural MT system should assign lower scores to divergent segments pairs than to those that are equivalent in meaning. We test this empirically using neural MT systems
trained for a single epoch, and use the system to score each of the sentence pairs in the test sets. We tune a threshold on the development set to convert scores to binary predictions. The system architecture and training settings are described in details in the later MT section (Section \ref{subsec:mtsystem}). Preliminary experiments showed that training for more than one epoch does not help divergence detection.

\subsection{Bilingual Sentence Embeddings}
\label{subsubsec:baseembed}

Our semantic divergence model introduces a large number of parameters to combine the pairwise word comparisons into a single sentence-level prediction.  This baseline tests whether a simpler model would suffice. We detect semantic divergence by computing the cosine similarity between sentence embeddings in a bilingual space. The sentence embeddings are bag-of-word representations, build by taking the mean of bilingual word embeddings for each word in the sentence.
This approach has been shown to be effective,
 despite ignoring fundamental linguistic
information such as word order and syntax \cite{MitchellLapata2010}. We use the same 200 dimensional BiVec word embeddings \cite{LuongPhamManning2015}, trained on OpenSubtitles and CommonCrawl respectively. 

\subsection{Textual Entailment Classifier}
\label{subsubsec:baseclte}

 Our final baseline replicates our previous system \cite{CarpuatVyasNiu2017} where we repurposed annotations and models designed for the task of Cross-Lingual Textual Entailment (CLTE) to detect semantic divergences. This baseline
also helps us understand how the synthetic training data compares to training examples generated manually, for a related cross-lingual  task.
Using CLTE datasets from SemEval \cite{NegriMarchettiMehdadBentivogliGiampiccolo2012,NegriMarchettiMehdadBentivogliGiampiccolo2013}, we train a supervised linear classifier that can distinguish sentence pairs that entail each other, from pairs where entailment does not hold in at least one direction. The features of the classifier are based on word alignments and sentence lengths.\footnote{As in the prior work, alignments are trained on 2M sentence pairs from Europarl \cite{Koehn2005} using the Berkeley aligner \cite{LiangTaskarKlein2006}. The classifier is the linear SVM from Scikit-Learn.}
\begin{table*}[t]
	\footnotesize
	\centering
	\begin{tabular}{lccccccc|ccccccc}
		\toprule	
		\textbf{Divergence Detection} & 		 \multicolumn{7}{c}{\textbf{OpenSubtitles}}  & \multicolumn{7}{c}{\textbf{Common Crawl}} \\ 
		\textbf{Approach} & +P &  +R &  +F &  -P &  -R&  -F& Overall F & +P &  +R &  +F &  -P &  -R&  -F & Overall F \\
		
		\midrule
		Sentence Embeddings &65 & 60 & 62 & 56 & 61 & 58 & 60  & 78 & 58 & 66 & 52 & \textbf{74} & 61 & 64\\
		MT Scores (1 epoch) & 67 & 53 & 59 & 54 & 68 & 60 & 60 & 54 & 65 & 59 & 17 & 11 & 14 & 42 \\ 
		Non-entailment & 58 & 78 & 66 & 53 & 30 & 38 & 54 &73 & 49 &58 & 48 & 72 & 57 & 58\\
		Non-parallel &  70 & 83 & 76 & 61 & 42 & 50 & 66 & 70 & 83 & 76 & 61 & 42 & 49 &  67 \\
		Semantic Dissimilarity & \textbf{76} & \textbf{80} & \textbf{78} & \textbf{75} & \textbf{70}& \textbf{72} & \textbf{77} & \textbf{82} &\textbf{ 88} &\textbf{ 85} &\textbf{ 78} & 69 & \textbf{73} & \textbf{80}\\		
		\bottomrule
	\end{tabular}
	\caption{Intrinsic evaluation on crowdsourced semantic equivalence vs. divergence testsets. We report overall F-score, as well as precision (P), recall (R) and F-score (F) for the equivalent (+) and divergent (-) classes separately. Semantic similarity yields better results across the board, with larger improvements on  the divergent class.}
	\label{tab:intrinsic}
	\vspace{-0.1in}
\end{table*}

\subsection{Intrinsic Evaluation Results}

 Table \ref{tab:intrinsic} shows that the semantic similarity model is most successful at distinguishing equivalent from divergent examples. The break down per class shows that both equivalent and divergent examples are better detected. The improvement is larger for divergent examples with gains of about 10 points for F-score for the divergent class, when compared to the next-best scores.

Among the baseline methods, the non-entailment model is the weakest. While it uses manually constructed training examples, these examples are drawn from distant domains, and the categories annotated do not exactly match the task at hand. In contrast, the other models benefit from training on examples drawn from the same corpus as each test set.

Next, the machine translation based model and the sentence embedding model, both of which are unsupervised, are significantly weaker than the two supervised models trained on synthetic data, highlighting the benefits of the automatically constructed divergence examples.  The strength of the semantic similarity model compared to the sentence embeddings model highlights the benefits of the fine-grained representation of bilingual sentence pairs as a similarity cube, as opposed to the  bag-of-words sentence embedding representation.

Finally, despite training on the same noisy synthetic data as the parallel v/s non-parallel system, the semantic similarity model is better able to detect meaning divergences. This highlights the benefits of (1) meaning comparison between words in a shared embedding space, over the discrete translation dictionary used by the baseline, and of (2) the deep convolutional neural network which enables the semantic comparison of overlapping spans in sentence pairs, as opposed to more local word alignment features.  

\subsection{Analysis}

We manually examine the 13-15\% of examples in each test set that are correctly detected as divergent by semantic similarity and mis-classified by the non-parallel detector.

On OpenSubtitles, most of these examples are true divergences rather than noisy alignments (i.e. sentences that are not translation of each other.) The non-parallel detector weighs length features highly, and is fooled by sentence pairs of similar length that share little content and therefore have very sparse word alignments. The remaining sentence pairs are plausible translations in some context that still contain inherent divergences, such as details missing or added in one language. The non-parallel detector views these pairs as non-divergent since most words can be aligned. The semantic similarity model can identify subtle meaning differences, and correctly classify them as divergent. As a result, the non-parallel detector has a higher false positive rate (22\%) than the semantic similarity classifier (14\%), while having similar false negative rates (11\% v/s 12\%).

On the CommonCrawl test set, the examples with disagreement are more diverse, ranging from noisy segments that should not be aligned to sentences with subtle divergences.

\section{Machine Translation Evaluation}
\label{sec:extrinsic}
Having established the effectiveness of the semantic divergence detector, we now measure the impact of divergences on a downstream  task, machine translation. As in our prior work \cite{CarpuatVyasNiu2017}, we take a data selection approach, selecting the least divergent examples in a parallel corpus based on a range of divergence detectors, and comparing the translation quality of the resulting neural MT systems. 

\subsection{Translation Tasks}

\paragraph{English-French} We evaluate on 4867 sentences from the Microsoft Spoken Language Translation dataset \cite{FedermannLewis2016} as well as on 1300 sentences from TED talks \cite{CettoloGirardiFederico2012}, as in past work \cite{CarpuatVyasNiu2017}. Training examples are drawn from  OpenSubtitles, which contains \texttildelow28M examples after deduplication. We discard 50\% examples for data selection.

\paragraph{Vietnamese-English} Since the \textsc{Semantic Similarity} model was designed to be easily portable to new language pairs, we also test its impact on the IWSLT Vietnamese-English TED  task, which comes with \texttildelow120,000 and 1268 in-domain sentences for training and testing respectively \citep{FarajianChatterjeeConfortiJalalvandBalaramanDiGangiAtamanTurchiNegriFederico2016}. This is a more challenging translation task as Vietnamese and English are more distant languages, there is little training data, and the sentence pairs are expected to be cleaner and more parallel than those from OpenSubtitles. In these lower resource settings, we discard 10\% of examples for data selection.

\subsection{Neural MT System}
\label{subsec:mtsystem}

We use the attentional encoder-decoder model \cite{BahdanauChoBengio2015} implemented in the SockEye toolkit \cite{HieberDomhanDenkowskiVilarSokolovCliftonPost2017}. Encoders and decoders are single-layer GRUs \cite{ChoVanmerrienboerGulcehreBahdanauBougaresSchwenkBengio2014} with 1000 hidden units. Source and target word embeddings have size 512. Using byte-pair encoding \cite{SennrichHaddowBirch2016b}, the vocabulary size is 50000. Maximum sequence length is set to 50.

We optimize the standard cross-entropy loss using Adam \cite{KingmaBa2014}, until validation perplexity does not decrease for 8 checkpoints. The learning rate is set to 0.0003 and is halved when the validation perplexity does not decrease for 3 checkpoints. The batch size is set to 80.  At decoding time, we construct a new model by averaging the parameters for the 8 checkpoints with best validation perplexity, and decode with a beam of 5. All experiments are run thrice with distinct random seeds. 

Note that the baseline in this work is much stronger than in our prior work ( $>$5 BLEU). This is due to multiple factors that have been recommended as best practices for neural MT and have been incorporated in the present baseline \---\ deduplication of the training data, ensemble decoding using multiple random runs, use of Adam as the optimizer instead of AdaDelta \cite{BaharAlkhouliPeterBrixNey2017,DenkowskiNeubig2017}, and checkpoint averaging \cite{BaharAlkhouliPeterBrixNey2017} \---\ as well as a more recent  neural modeling toolkit.

\subsection{English-French Results}

We train English-French neural MT systems by selecting the least divergent half of the training corpus with the following criteria:
\begin{itemize}[noitemsep]
	\item \textsc{Semantic Similarity} (Section~\ref{sec:model})
	\item \textsc{Parallel}: the non-parallel sentence detector (Section~\ref{subsubsec:basemm})
	\item \textsc{Entailment}: the entailment classifier  (Section~\ref{subsubsec:baseclte}), as in  \citet{CarpuatVyasNiu2017} 
		\item \textsc{Random}: Randomly downsampling the training corpus
\end{itemize}

\begin{figure*}[htbp]
		\centering
		\includegraphics[width=6in,trim={0 9.7cm 0 3cm},clip]{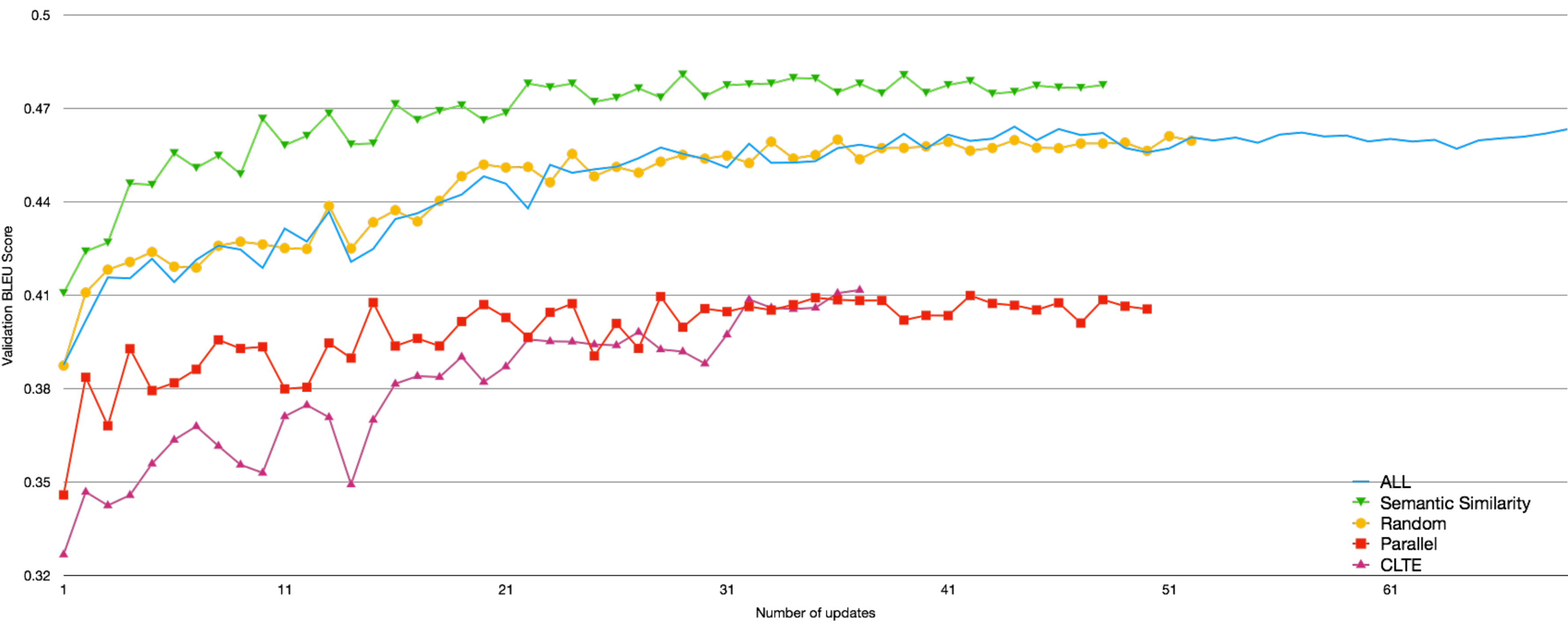}
		\vspace{-0.1in}
		\caption{Learning curves on the validation set for English-French models (mean of 3 runs/model). The \textsc{Semantic Similarity} model outperforms other models throughout training, including the one trained on all data.}
		\label{fig:learningcurves}
		\vspace{-0.05in}
	\end{figure*}

\begin{table}[t]

	\small
	\centering
	\begin{tabular}{lcc|cc}
		\toprule	
		\multicolumn{1}{c}{\textbf{Model }}
		& \multicolumn{2}{c}{\textbf{MSLT BLEU}} &  \multicolumn{2}{c}{\textbf{TED BLEU}} \\
		& Avg. & Ensemble &  Avg. & Ensemble \\
		
		\midrule
		\textsc{Random}  &  43.49 & 45.64 & 36.05 & 38.20\\
		\textsc{Parallel} &  40.65 & 42.12 &  35.99 &  37.86\\
		\textsc{Entailment} &  39.64 & 41.86 &  33.30 & 35.40\\
        \textsc{Semantic Sim.}   & \textbf{45.53} & \textbf{47.23}*  & \textbf{36.98} & \textbf{38.87} \\
		\midrule
\textsc{All} &  44.64 & 46.26  &  36.98 & 38.59\\

		\bottomrule
		
	\end{tabular}
	\caption{English-French decoding results. BLEU scores are either averaged across 3 runs (``Avg.") or obtained via ensemble decoding (``Ensemble"). \textsc{Semantic Similarity} reaches BLEU scores on par with \textsc{All} with only half of the training data. \textsc{Semantic Similarity} scores marked with * are significanly better (p $<$ 0.05) than the corresponding \textsc{All} scores.}
	\label{tab:fren}
	\vspace{-0.16in}
\end{table}

Learning curves (Figure \ref{fig:learningcurves}) show that data selected using \textsc{Semantic Similarity} yields better validation BLEU throughout training compared to all other models. \textsc{Semantic Similarity} selects more useful examples for MT than \textsc{Parallel}, even though both selection models are trained on the same synthetic examples. This highlights the benefits of semantic modeling over surface mis-alignment features. Furthermore, \textsc{Semantic Similarity} achieves the final validation BLEU of the model that uses \textsc{All} data with only 30\% of the updates. This suggests that semantically divergent examples are pervasive in the training corpus, confirming the findings from  manual annotation  (Section \ref{sec:manual}), and that the presence of such examples slows down neural MT training.

Decoding results on the blind test sets (Table \ref{tab:fren}) show that   \textsc{Semantic Similarity} outperforms all other data selection criteria  (with differences being statistically significant (p $<$ 0.05) \cite{Koehn2004stat}), and performs as well or better than the \textsc{All} model which has access to twice as many training examples.

The \textsc{Semantic Similarity} model also achieves significantly better translation quality than the \textsc{Entailment} model used in our prior work. Surprisingly, the \textsc{Entailment} model performs worse than the \textsc{All} baseline, unlike in our prior work. We attribute this different behavior to several factors:  the strength of the new baseline (Section \ref{subsec:mtsystem}), the use of Adam instead of AdaDelta, which results in stronger BLEU scores at the beginning of the learning curves for all models, and finally the deduplication of the training data. In our prior systems, the training corpus was not deduplicated. Data selection had a side-effect of reducing the ratio of duplicated examples. When the \textsc{Entailment} model was used, longer sentence pairs with more balanced length were selected, yielding longer translations with a better BLEU brevity penalty than the baseline.  With the new systems, these advantages vanish.  We further analyze output lengths in Section \ref{subsec:analysis}.

\subsection{Vietnamese-English Results}

\begin{table}[t]
	\footnotesize
	\centering
	\begin{tabular}{lc}
		\toprule	
		\multicolumn{1}{l}{\textbf{Model}} & \multicolumn{1}{c}{\textbf{Avg. Test Set BLEU}} \\
		\midrule
		\textsc{Random (90\%)}   & 22.71 \\
		\textsc{Semantic Sim. (90\%)}   & \textbf{23.38} \\

		\midrule
\textsc{All}  & 23.30\\
		\bottomrule
		
	\end{tabular}
	\caption{Vietnamese-English decoding results: dropping 10\% of the data based on \textsc{Semantic Similarity} does not hurt BLEU, and performs significantly (p $<$ 0.05) better than \textsc{Random} selection.}
	\label{tab:vien}
		\vspace{-0.1in}
\end{table}

Trends from English-French carry over to Vietnamese English, as the \textsc{Semantic Similarity} model compares favorably to \textsc{All} while reducing the number of training updates by 10\%. \textsc{Semantic Similarity} also yields better BLEU than \textsc{Random} with the differences being statistically significant.  While differences in score here are smaller, these result are encouraging since they demonstrate that our semantic divergence models port to more distant low-resource language pairs.

 \subsection{Analysis}
\label{subsec:analysis}

\begin{figure}[htbp]
 	 		\centering
 	 		\includegraphics[width=3in]{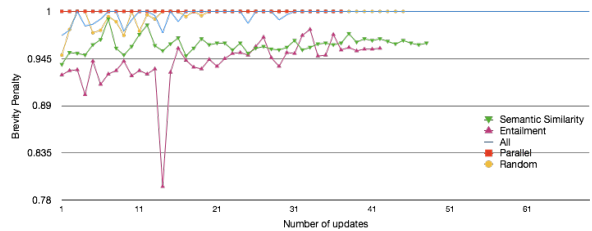}
 	 		\caption{Brevity penalties on the validation set for English-French models (single run). }
 	 		\label{fig:bpcurves}
 	 			\vspace{-0.1in}
 	 	\end{figure}

 We break down  the results seen in Figure \ref{fig:learningcurves} and Table \ref{tab:fren}, with a focus on the behavior of the \textsc{Entailment} and \textsc{All} models. We start by analyzing the BLEU brevity penalty trends observed on the validation set during training (Figure \ref{fig:bpcurves}). 	
 	
 We observe that both the \textsc{Entailment} and \textsc{Semantic Similarity} based models have similar brevity penalties despite having performances that are at opposite ends of the spectrum in terms of BLEU. This implies that translations generated by the \textsc{Semantic Similarity} model have better n-gram overlap with the reference, but are much shorter. Manual examination of the translations suggests that the \textsc{Entailment} model often fails by under-translating sentences, either dropping segments from the beginning or the end of source sentences (Table \ref{tab:analysis1}). 
 
 The \textsc{Parallel} model consistently produces translations that are longer than the reference. \footnote{The brevity penalty does not penalize translations that are longer than the reference.} This is partially due to the model's propensity to generate a sequence of garbage tokens in the beginning of a sentence, especially while translating shorter sentences. In our test set, almost 12\% of the translated sentences were found to begin with the garbage text shown in Table \ref{tab:analysis1}. Only a small fraction ($ < 0.02\%$) of the French sentences in our training data begin with these tokens, but the tendency of \textsc{Parallel} to promote divergent examples above non-divergent ones, seems to exaggerate the generation of this sequence.

   \begin{table}[t]\footnotesize
   	\begin{tabular}{lp{5.5cm}}
   		\toprule
		 \multicolumn{2}{c}{\textbf{\textsc{Entailment} is inadequate due to under-translation}} \\
		\midrule
   		Source &  he's a very impressive man  \textbf{and still goes out to do digs.} \\
   		Reference &  c'est un homme tr\`es impressionnant \textbf{et il fait encore des fouilles.} \\
   		\textsc{Entailment} & c'est un homme tr\`es impressionnant. \\
   		\midrule
   			Source &   when the Heat \textbf{first} won. \\
   			Reference &  lorsque les Heat ont gagn\'e \textbf{pour la premi\`ere fois}. \\
   			\textsc{Entailment} & quand le Heat a gagn\'e. \\
   		\midrule
   		 \multicolumn{2}{c}{\textbf{\textsc{Parallel} produces garbage tokens}} \\
   		\midrule
   			Source &   alright. \\
   			Reference &  	d'accord. \\
   			\textsc{Entailment} & \textbf{\{ $\backslash$ pos (192,210)\}} d'accord. \\
   	
   		\bottomrule
   	\end{tabular}
   	\caption{Selected translation examples from the ensemble systems of the various models. }
   	\label{tab:analysis1}
   	\vspace{-0.1in}
   \end{table}

\section{Conclusion}

We conducted an extensive empirical study of semantic divergences in parallel corpora. Our crowdsourced annotations confirms that correctly aligned sentences are not necessarily meaning equivalent. We introduced an approach based on neural semantic similarity that detects such divergences much more accurately than shallower translation or alignment based models. Importantly, our model does not require manual annotation, and can be trained for any language pair and domain with a parallel corpus. Finally, we show that filtering out divergent examples helps speed up the convergence of neural machine translation training without loss in translation quality, for two language pairs and data conditions. New datasets and models introduced in this work are available at \url{http://github.com/yogarshi/SemDiverge}.

These findings open several avenues for future work: How can we improve divergence detection further? Can we characterize the nature of the divergences beyond binary predictions? How do divergent examples impact other applications, including cross-lingual NLP applications and semantic models induced from parallel corpora, as well as tools for human translators and second language learners?
\section*{Acknowledgments}
{\small
We thank the CLIP lab at the University of Maryland and the anonymous reviewers from NAACL 2018 and WMT 2017 for their constructive feedback. This work was supported in part by research awards from Amazon, Google, and the Clare Boothe Luce Foundation.}

\bibliographystyle{acl_natbib}
\bibliography{zotero,zotero_yogarshi,refs}
\end{document}